\PassOptionsToPackage{table}{xcolor}
\documentclass[sigconf,nonacm]{acmart}

\usepackage{amsmath}
\usepackage{multirow}
\usepackage{booktabs}
\usepackage{pifont}
\AtBeginDocument{%
}

\usepackage{graphicx}
\usepackage{subcaption}
\usepackage{array}
\usepackage{enumitem}
\usepackage[ruled,vlined]{algorithm2e}

\newcommand{\cmark}{\ding{51}}
\newcommand{\xmark}{\ding{55}}
\newcommand{\up}[1]{%
  \textcolor{green!60!black}{\scriptsize +#1}%
}
\definecolor{ForestGreen}{RGB}{34,139,34}
\definecolor{MemoLinkBlue}{RGB}{95,141,187}

\begin{document}

\title{
MEMO: Multi-Level Entity-Aware Memory for\\
Streaming Video Understanding
}

\author{%
\begin{tabular}{c}
Yinying Li$^{1,*}$ \quad
Yuqian Fu$^{2,*}$ \quad
Yulin Dai$^{1,*}$ \quad
Jingyu Gong$^{1}$ \quad
Tianwen Qian$^{1,\dagger}$ \quad
Xiaoling Wang$^{1,\dagger}$
\\[3pt]
$^{1}$East China Normal University
\quad
$^{2}$King Abdullah University of Science and Technology
\end{tabular}
}

\renewcommand{\shortauthors}{Yinying Li et al.}

\begin{abstract}
Streaming video understanding requires models to process unbounded visual
streams while preserving rich visual semantics across vast temporal horizons,
posing a fundamental challenge for memory modeling. Existing approaches
primarily focus on increasing memory capacity, either by compressing historical
information into fixed-size representations or by extending storage beyond GPU
memory. However, these methods largely rely on global or coarse-grained
representations, inevitably losing fine-grained visual information. In this
work, we argue that streaming video memory should explicitly encode structured
and semantically meaningful representations, particularly at the entity level.
To this end, we propose \textbf{MEMO}, a novel framework that models streaming
video through multi-level, entity-aware structured memory. MEMO performs
multi-level perception to jointly capture global semantics, entity dynamics,
and spatial structures, partitioning streaming video into semantically coherent
chunks. Each chunk is organized into a structured memory, where lightweight
global and entity-level representations serve as retrieval indices, while the
corresponding high-resolution visual content is retained separately for
on-demand access. At inference time, MEMO performs query-specific retrieval
over the structured memory and selectively recalls relevant visual evidence for
downstream reasoning. Notably, MEMO is training-free and plug-and-play with
existing multimodal large language models. Extensive experiments on
StreamingBench and OVO-Bench demonstrate that MEMO consistently improves
multiple base models and achieves state-of-the-art performance. 
The code is available at \href{https://github.com/Cecilia384/MEMO}{\textcolor{MemoLinkBlue}{\nolinkurl{https://github.com/Cecilia384/MEMO}}}.
\end{abstract}

\maketitle

\begin{figure}[!t]
    \centering
    \includegraphics[width=\columnwidth]{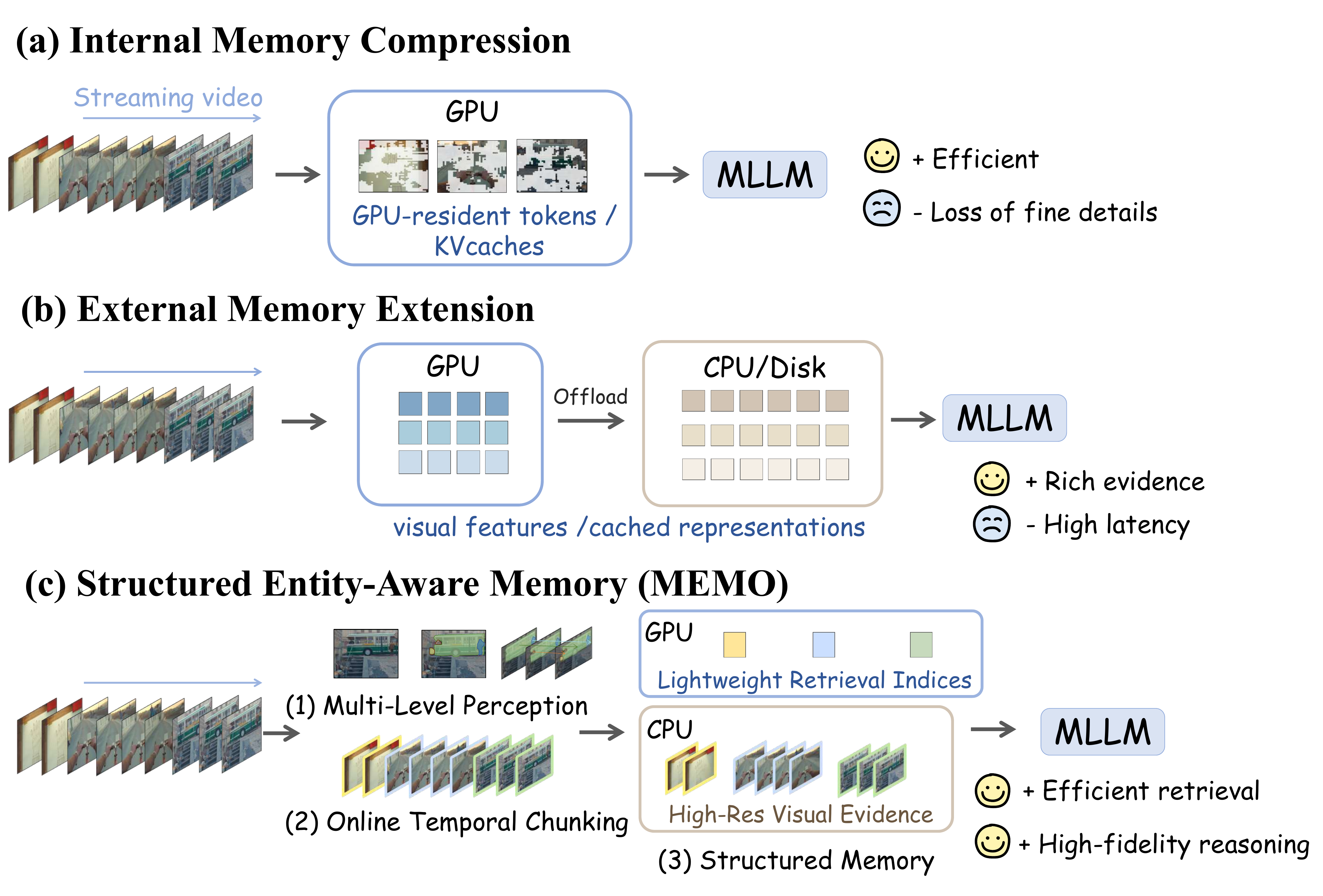}
    \caption{
    Comparison of memory paradigms for streaming video understanding.
    (a) Internal memory compression is efficient but loses fine-grained
    details. (b) External memory extension retains longer history but relies
    on coarse-grained evidence. (c) MEMO decouples lightweight structured
    indices from high-resolution visual evidence.
    }
    \Description{
    Comparison of internal memory compression, external memory extension,
    and MEMO's structured entity-aware memory.
    }
    \label{img:streaming-definition}
    \vspace{-1em}
\end{figure}

\section{Introduction}

The rapid advancement of Multimodal Large Language Models (MLLMs)~\cite{li2024llava,wang2022omnivl,bai2025qwen25vltechnicalreport,wang2025internvl3,bai2025qwen3} has significantly improved video understanding. However, most existing methods~\cite{lei2021less,shao2025tokens} rely on an unrealistic assumption that the entire video is pre-recorded and fully accessible at the onset of inference. In real-world scenarios such as embodied navigation~\cite{zheng2024towards,li2026bridging}, autonomous driving~\cite{chen2024end,brodermann2025cafuser}, live broadcasting~\cite{yang2025livestar}, and egocentric vision~\cite{plizzari2024outlook,li2026egocross,zhang2026egonight,zhu2026egosound}, video arrives as a continuous and unbounded stream. This fundamental mismatch motivates the need for streaming video understanding~\cite{chen2024videollm}.

A few recent works have begun to explore streaming video understanding from the perspective of memory modeling. As illustrated in Figure~\ref{img:streaming-definition}, early approaches primarily rely on internal memory mechanisms, where historical visual information is compressed into GPU-resident representations such as tokens or key-value caches~\cite{yao2025timechat, yang2025streammem}. While such a design enables efficient and low-latency processing, the limited capacity of GPU memory necessitates aggressive compression, which inevitably leads to information loss. To alleviate memory constraints, subsequent methods extend memory beyond the GPU by leveraging additional storage resources such as CPU or disk. Representative examples include ReKV~\cite{di2025streaming} and LiveVLM~\cite{ning2025livevlm}, which retain more historical information by offloading visual features or cached representations. However, despite improving storage capacity, these approaches largely focus on preserving global or coarse-grained visual information, without explicitly modeling finer-grained semantic structures.

We argue that effective streaming video understanding requires more than simply increasing memory capacity. To support accurate reasoning over evolving streams, the memory must preserve structured and semantically meaningful representations rather than coarse, weakly organized historical traces. A key step in this direction is to explicitly model entity-level information, since objects constitute the primary carriers of scene semantics and their dynamics provide strong cues for boundaries in event evolution.

Building on this insight, we propose \textbf{MEMO}, a novel framework that organizes streaming video into \emph{multi-level, entity-aware structured memory}. Rather than relying on coarse or heavily compressed representations, MEMO performs multi-level perception that jointly models global semantics, entity dynamics, and spatial structures. These cues enable robust estimation of temporal continuity, allowing unbounded streams to be dynamically segmented into semantically coherent chunks. Each chunk is represented by global summaries and fine-grained entity features, forming a structured memory that preserves scene context and detailed object-level information. To further support efficient and scalable storage, MEMO adopts a decoupled memory design, where lightweight representations serve as retrieval indices, while high-resolution visual evidence is stored separately for on-demand access. At inference time, MEMO performs query-specific retrieval over the structured memory to identify relevant chunks and selectively recalls their associated visual evidence for reasoning. 
Importantly, MEMO is \textit{training-free} and \textit{plug-and-play}, enabling seamless integration with any MLLM without additional training or optimization.

Extensive experiments demonstrate that MEMO consistently improves performance across multiple base models (e.g., LLaVA-OV-series~\cite{li2024llava}, Qwen-VL-series~\cite{bai2025qwen25vltechnicalreport, bai2025qwen3}) and benchmarks (OVO-Bench~\cite{niu2025ovo}, StreamingBench~\cite{lin2024streamingbench}). Notably, MEMO not only enhances open-source base models but also achieves competitive or even superior performance compared to strong training-based and proprietary systems. For example, when applied to Qwen3-VL-8B, MEMO improves the baseline by 10.5\% and surpasses Gemini 1.5 Pro and GPT-based systems by 8.00\% and 10.41\%, respectively.

In summary, our main contributions are three-fold:
\begin{itemize}
     \item We propose MEMO, a novel framework that models streaming video through multi-level, entity-aware structured memory, addressing the limitations of existing memory representations for unbounded visual streams.
    \item We propose a multi-level perception and structuring mechanism that models global semantics, entity dynamics, and spatial structures, enabling robust temporal segmentation into coherent, entity-aware chunks.
    \item Extensive experiments on StreamingBench and OVO-Bench demonstrate that MEMO consistently outperforms existing methods and achieves state-of-the-art performance. 
\end{itemize}

\section{Related Work}

\noindent
\textbf{MLLMs for Video Understanding.}
Recent multimodal large language models (MLLMs) have achieved remarkable progress in visual understanding by integrating visual perception with language reasoning, demonstrating strong generalization across a wide range of multimodal applications, including visual question answering~\cite{liu2023visual,li2025videochat,balauca2025understanding}, image captioning~\cite{bai2025qwen25vltechnicalreport}, and visual grounding tasks such as detection and segmentation~\cite{oquab2023dinov2,ravi2024sam,fu2025objectrelator,pan2026v}. Flagship ones include LLaVA-OneVision-1.5~\cite{an2025llava}, Qwen3-VL~\cite{bai2025qwen3}, InternVL3.5~\cite{wang2025internvl3}, VideoLLaMA3~\cite{zhang2025videollama}. Building upon this success, many MLLMs naturally extend to the video domain, e.g., Video-ChatGPT~\cite{maaz2024video}, LLaVA-NeXT~\cite{liu2024llavanext}, by processing sampled frames or short clips as sequential visual inputs, enabling joint reasoning over temporal visual tokens. In addition, several recent works~\cite{song2024moviechat} are specifically designed for video understanding, incorporating temporal modeling or long-context mechanisms to better capture dynamic visual content. Despite these advances, most existing approaches still operate under an offline paradigm, assuming that the entire video is pre-recorded and fully accessible during inference. This assumption fundamentally limits their applicability in streaming scenarios, where video arrives as a continuous and unbounded sequence and requires incremental, real-time processing.

\noindent
\textbf{Streaming Video Understanding.}
Recent efforts have explored streaming video understanding to enable real-time processing over continuous visual streams~\cite{huang2025online,chen2024videollm,xiong2025streaming,liu2024streamchat,qian2024streaming, wang2026streameqa,dong2026objectstream}. 
A line of work adopts internal memory mechanisms, where historical visual information is maintained in compressed GPU-resident representations such as tokens or key-value caches. Methods such as Flash-VStream~\cite{zhang2025flash} and StreamMem~\cite{yang2025streammem} follow this design to support efficient sequential processing. To further mitigate computational overhead during real-time inference, TimeChat-Online ~\cite{yao2025timechat} prunes visual redundancies via dynamic token dropping and hierarchical compression, similarly, FluxMem~\cite{xie2026fluxmem} introduces a training-free adaptive hierarchical memory to progressively reduce spatiotemporal token redundancy based on intrinsic scene statistics, while StreamingVLM ~\cite{xu2025streamingvlm} stabilizes generation by maintaining a compact, asymmetric KV cache with attention sinks and sliding windows.
Another line of work extends memory beyond the GPU by leveraging external storage (e.g., CPU or disk) to retain richer historical information and retrieve relevant context on demand. Representative methods include ReKV~\cite{di2025streaming}, which retrieves historical KV caches from RAM/disk, LiveVLM~\cite{ning2025livevlm}, which further combines compressed KV memory with short- and long-term retrieval, and Vista~\cite{lu2026vista}, which organizes streaming history as scene-level memory for selective recall. However, these methods mostly retrieve coarse-grained context units, with limited explicit modeling of fine-grained entity structures. Our method instead constructs a multi-level, entity-aware structured memory that explicitly encodes both global context and entity dynamics.

\section{Methodology}

\begin{figure*}[t]
    \centering
    \includegraphics[width=\textwidth]{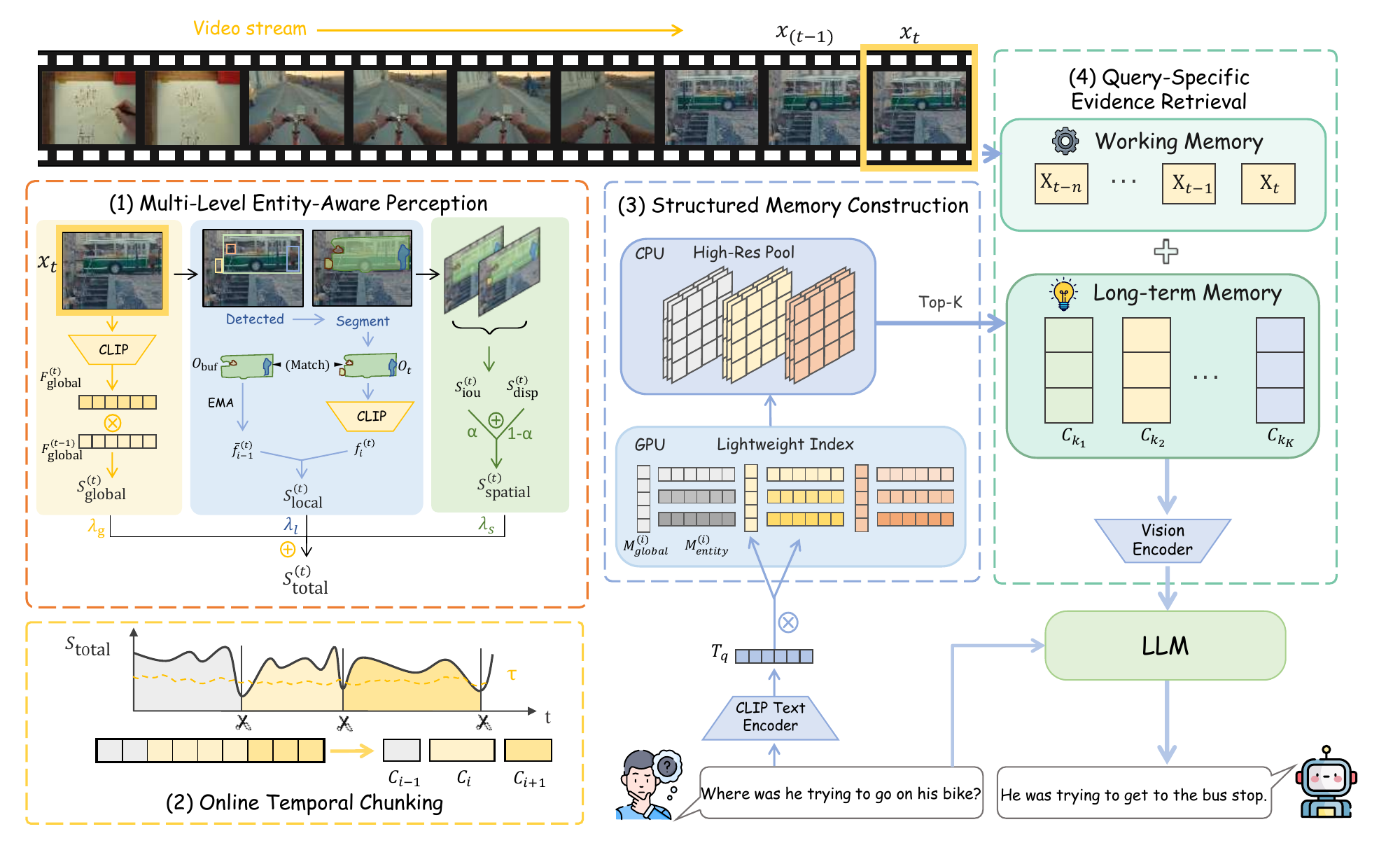}
    \caption{
    \textbf{Overall architecture of MEMO.}
    The streaming pipeline consists of four stages:
    (1) \textbf{Multi-Level Entity-Aware Perception} extracts global,
    entity-level, and spatial cues to model semantic continuity;
    (2) \textbf{Online Temporal Chunking} incrementally partitions the
    video stream into semantically coherent chunks based on similarity;
    (3) \textbf{Structured Memory Construction} organizes chunks into
    a structured memory with lightweight indices and high-resolution
    visual evidence; and
    (4) \textbf{Query-Specific Evidence Retrieval} retrieves the most
    relevant chunks and their associated visual evidence to support
    downstream reasoning.
    }
    \Description{
    A left-to-right overview of the MEMO pipeline begins with a
    continuous video stream. Multi-level perception combines
    whole-frame CLIP similarity, matched entity features updated by
    exponential moving average, and spatial cues based on mask overlap
    and object displacement into a total similarity score. Online
    temporal chunking compares this score with an adaptive threshold to
    partition the stream into semantic chunks. For each chunk,
    high-resolution visual evidence is stored in a CPU pool, while
    lightweight global and entity indices remain on the GPU. At query
    time, the text representation retrieves the top-K historical chunks.
    Their visual evidence is combined with frames from the current
    working memory, encoded, and passed with the query to the language
    model to generate an answer.
    }
    \label{fig:framework}
\end{figure*}

\noindent
\textbf{Problem Setup.}
We consider streaming video understanding under a strict online setting, where video frames arrive sequentially over time. At each time step $t$, the model receives the current frame $x_t$ and performs reasoning based solely on $x_t$ and past observations $\{x_1, \dots, x_{t-1}\}$, without access to future frames.

Given a query $Q$ at time $t$, the goal of the MLLM $\phi$ is to produce an answer $A_t$ by leveraging both the current frame and relevant historical visual information. This requires the model to maintain and continuously update an internal representation of past observations as new frames arrive.

\noindent
\textbf{Method Overview.}
We address streaming video understanding by constructing a multi-level, entity-aware structured memory that supports accurate reasoning over unbounded visual streams. As illustrated in Figure~\ref{fig:framework}, our proposed MEMO framework consists of four key components: (1) \textit{Multi-Level Entity-Aware Perception}, (2) \textit{Online Temporal Chunking}, (3) \textit{Structured Memory Construction}, and (4) \textit{Query-Specific Evidence Retrieval}.

Overall, given a video stream $\{x_1, \dots, x_t\}$, MEMO first performs \textit{multi-level entity-aware perception} to extract complementary visual cues, including global semantic context, local entity features, and spatial structures. Based on these representations, MEMO computes a unified similarity score between the current frame $x_t$ and its historical context, capturing semantic continuity from both global and fine-grained levels. This score guides \textit{online temporal chunking} to partition the stream into semantically coherent chunks. The resulting chunks are organized by the \textit{structured memory construction} module into a structured memory, where lightweight multi-level representations serve as retrieval indices and high-resolution visual evidence is preserved separately for on-demand access.
Given a query $Q$, the \textit{query-specific evidence retrieval} module selects the most relevant chunks and retrieves their associated visual evidence for downstream reasoning.

\subsection{Multi-Level Entity-Aware Perception}
\label{sec:method_perception}

To handle continuous and unbounded visual streams, MEMO models semantic continuity between the current frame and its recent context through multi-level, entity-aware visual representations. Specifically, we assess semantic consistency from three complementary perspectives: \emph{global semantics}, \emph{local object semantics}, and \emph{spatial structure}.

\noindent
\textbf{Global semantic similarity.}
We first define the global semantic similarity $S_{\mathrm{global}}^{(t)}$ to measure the semantic continuity between two adjacent frames at the whole-frame level. We use CLIP~\cite{radford2021learning} to extract global semantic features for each frame. Let $F_{\mathrm{global}}^{(t-1)}, F_{\mathrm{global}}^{(t)} \in \mathbb{R}^{D}$ denote the global feature vectors of frames $x_{t-1}$ and $x_t$, respectively, where $D$ is the feature dimension. After $L_2$ normalization, we compute the cosine similarity as:
\begin{equation}
S_{\mathrm{global}}^{(t)}
=
\frac{1 + \cos\!\big(F_{\mathrm{global}}^{(t-1)}, F_{\mathrm{global}}^{(t)}\big)}{2}.
\end{equation}

Although global frame-level features capture overall scene semantics, they are often insufficient to characterize changes in key semantic entities. Therefore, we further incorporate entity-level modeling to measure similarity from the perspective of tracked objects.
We first employ Grounding DINO~\cite{liu2024grounding} for object detection and SAM~\cite{kirillov2023segment} for segmentation, and then associate objects across frames to assign stable identities. Let $\mathcal{O}_t$ denote the set of objects in frame $x_t$. For each persistently tracked object, we maintain a historical semantic state in memory. On top of this memory, we define two object-level measures: local semantic continuity $S_{\mathrm{local}}^{(t)}$ and spatial structural continuity $S_{\mathrm{spatial}}^{(t)}$.

\noindent
\textbf{Local semantic continuity.}
The local semantic similarity $S_{\mathrm{local}}^{(t)}$ measures the temporal consistency of object appearance. For each object $i$, we suppress the background using the segmentation mask generated by SAM and extract an object-level CLIP feature, denoted by $f_i^{(t)}$. To reduce semantic fluctuations caused by single-frame noise, we maintain a historical semantic state $\bar{f}_i^{(t)}$ for each object and update it with exponential moving average (EMA):
\begin{equation}
\bar{f}_i^{(t)}=\rho f_i^{(t)}+(1-\rho)\bar{f}_i^{(t-1)},
\end{equation}
where $\rho$ is the EMA update coefficient.
For objects that can be successfully matched with historical states, we compute the cosine similarity between the current observation and its historical semantic state:
\begin{equation}
S_{\mathrm{local},i}^{(t)}
=
\frac{1 + \cos\!\big(f_i^{(t)}, \bar{f}_i^{(t-1)}\big)}{2}.
\end{equation}

Considering that different objects may contribute unequally to scene semantics, we aggregate object-level scores using object weights $w_i$, which denote the pixel-area ratio of object $i$. Thus, $S_{\mathrm{local}}^{(t)}$ is defined as:
\begin{equation}
S_{\mathrm{local}}^{(t)}
=
\frac{\sum_{i \in \mathcal{O}_t \cap \mathcal{O}_{\mathrm{buf}}} w_i S_{\mathrm{local},i}^{(t)}}
{\sum_{i \in \mathcal{O}_t \cap \mathcal{O}_{\mathrm{buf}}} w_i}
-
P_{\mathrm{penalty}}^{(t)},
\end{equation}
where $\mathcal{O}_{\mathrm{buf}}$ denotes the set of objects stored in the historical buffer, and $\mathcal{O}_t \cap \mathcal{O}_{\mathrm{buf}}$ denotes the set of objects that are both currently visible and successfully aligned with historical memory. The penalty term $P_{\mathrm{penalty}}^{(t)}$ is introduced to account for object-set changes between the current frame and the buffered history.

We formulate a discontinuity penalty to account for both newly appeared objects, defined as $\mathcal{O}_{\text{new}} = \mathcal{O}_t \setminus \mathcal{O}_{\mathrm{buf}}$, and disappeared objects, defined as $\mathcal{O}_{\text{dis}} = \mathcal{O}_{\mathrm{buf}} \setminus \mathcal{O}_t$. Intuitively, the emergence of new objects often indicates semantic shifts (e.g., shot transitions), whereas object disappearance is more commonly caused by transient factors such as occlusions. To capture this asymmetry, we assign different penalty weights $\gamma_{\text{new}} > \gamma_{\text{dis}}$, imposing a stronger penalty on newly appeared objects.

Specifically, the penalty for newly appeared objects is defined as the relative weight ratio of new objects in the current frame. In contrast, temporarily disappeared objects are maintained in a hidden state with a dedicated counter $c_k$. To enhance robustness against occlusions and tracking jitter, we use a bounded quadratic function
$g(c_k)=(\min(c_k/H_{\mathrm{hid}},1))^2$ to progressively increase
the penalty with the hidden duration. The overall penalty term is defined as:
\begin{equation}
P_{\mathrm{penalty}}^{(t)} = \gamma_{\mathrm{new}} \left( \frac{\sum_{i \in \mathcal{O}_{\text{new}}} w_i}{\sum_{j \in \mathcal{O}_t} w_j} \right) + \gamma_{\mathrm{dis}} \sum_{k \in \mathcal{O}_{\text{dis}}} w_k \cdot g(c_k).
\end{equation}
where $w_i$, $w_j$, and $w_k$ denote the respective target's area proportion relative to the entire frame, and $\mathcal{O}_t$ represents the complete set of valid objects in the current frame.
These two terms explicitly quantify the discontinuity induced by the emergence of new objects and the prolonged absence of existing ones, respectively. As either term increases, the local semantic continuity score decreases accordingly.

\noindent
\textbf{Spatial structural continuity.}
The spatial similarity $S_{\mathrm{spatial}}^{(t)}$ focuses on geometric continuity between matched objects across adjacent frames. For each successfully matched object, we describe its spatial consistency from two aspects. First, we compare the overlap of its segmented regions in two adjacent frames and obtain the mask intersection-over-union score $S_{\mathrm{iou}}^{(t)}$. Second, we compare the displacement of the object bounding-box center between two frames and obtain the displacement consistency score $S_{\mathrm{disp}}^{(t)}$. The final spatial continuity is defined as:
\begin{equation}
S_{\mathrm{spatial}}^{(t)}
=
\alpha S_{\mathrm{iou}}^{(t)} + (1-\alpha) S_{\mathrm{disp}}^{(t)},
\end{equation}

where $\alpha$ balances mask overlap and displacement smoothness.
Specifically, $S_{\mathrm{iou}}^{(t)}$ is the average mask IoU over
matched objects, while $S_{\mathrm{disp}}^{(t)}$ averages their
displacement consistency, computed from the normalized bounding-box
center displacement. If no object is matched, both scores are set to
zero. We set $\alpha=0.6$ and the maximum displacement tolerance to
$d_{\max}=0.5$.

\subsection{Online Temporal Chunking}
\label{sec:method_segmentation}

After computing the three similarity components, we combine them into an overall similarity score:
\begin{equation}
S_{\mathrm{total}}^{(t)}
=
\lambda_s S_{\mathrm{spatial}}^{(t)}
+
\lambda_l S_{\mathrm{local}}^{(t)}
+
\lambda_g S_{\mathrm{global}}^{(t)},
\end{equation}
where $\lambda_s$, $\lambda_l$, and $\lambda_g$ are hyperparameters.

When a new frame $x_t$ arrives, we compute its corresponding $S_{\mathrm{total}}^{(t)}$ and incorporate it into the currently growing segment. To this end, we maintain an \emph{active segment buffer}, which stores the continuous frame sequence of the current unfinished segment. The buffer grows as new frames arrive until the system determines that a semantic boundary has been reached.
Meanwhile, we maintain a historical statistics window to record recent total similarity scores $S_{\mathrm{total}}^{(t-n)}, \dots, S_{\mathrm{total}}^{(t)}$, where $n$ is the window size. Based on the mean $\mu^{(t)}$ and standard deviation $\sigma^{(t)}$ within this window, we estimate an adaptive threshold,

\begin{equation}
\tau^{(t)}
=
\mathrm{clip} (\mu^{(t)} - m \cdot \sigma^{(t)}, \tau_{\min}, \tau_{\max}),
\end{equation}
where $m$ controls the sensitivity to similarity drops, 
and $\mathrm{clip}(\cdot)$ denotes a truncation function that restricts the threshold to lie within $[\tau_{\min}, \tau_{\max}]$.

Because motion blur, short-term occlusion, and local perception noise
may cause transient fluctuations in the similarity score, a single
low-similarity observation may not reliably indicate a semantic boundary.
We therefore confirm a boundary only when
$S_{\mathrm{total}}^{(t)}<\tau^{(t)}$ persists for
$N_{\mathrm{confirm}}$ consecutive frames. To further avoid
over-segmentation and fragmented chunks, we impose a minimum
segment-length constraint. A cut is triggered only when both the
low-similarity confirmation and the minimum-length requirement are
satisfied.
 
\subsection{Structured Memory Construction}
\label{sec:method_storage}

For any segmented semantic chunk, we adopt a hierarchical storage strategy:

\noindent
\noindent
\textbf{High-resolution visual evidence.}
For each chunk $C^{(i)}$, we store the original high-resolution visual tensor $X^{(i)} \in \mathbb{R}^{T_i \times H \times W \times 3}$ in CPU memory as visual evidence for downstream fine-grained reasoning. The visual evidence is encoded on demand: only frames sampled from retrieved chunks are processed by the vision encoder of the MLLM $\phi$, rather than encoding the entire historical memory. For multi-turn queries, MLLM-specific visual representations are cached after first use and reused when the same evidence is retrieved again.

\noindent
\textbf{Lightweight structured memory.}
On the GPU side, we maintain only compact chunk-level memory: $\mathcal{M}^{(i)} = \langle \mathcal{M}_{\mathrm{global}}^{(i)},\; \mathcal{M}_{\mathrm{entity}}^{(i)} \rangle$,

which serve as efficient indices for similarity-based retrieval. Specifically, $\mathcal{M}_{\mathrm{global}}^{(i)}$ denotes the global semantic representation of chunk $C^{(i)}$, which is defined as:
\begin{equation}
\mathcal{M}_{\mathrm{global}}^{(i)} =
\frac{\sum_{j=1}^{T_i} \omega_j F_{\mathrm{global}}^{(j)}}
{\left\|\sum_{j=1}^{T_i} \omega_j F_{\mathrm{global}}^{(j)}\right\|_2}
\in \mathbb{R}^D,
\end{equation}
where $F_{\mathrm{global}}^{(j)} \in \mathbb{R}^D$ denotes the global feature of the $j$-th frame in the current chunk $C^{(i)}$, and $\omega_j  \propto \max(c, 1 - S_{\mathrm{total}}^{(j-1)})$ is the aggregation weight with a constant $c$.

For the first frame in each chunk, we set $\omega_1 = 1$.
Correspondingly,  $\mathcal{M}_{\mathrm{entity}}^{(i)}$ denotes the  entity-level representation of chunk $C^{(i)}$, which is defined as:

\begin{equation}
\mathcal{M}_{\mathrm{entity}}^{(i)} = [\bar{f}_1^{(i)}, \bar{f}_2^{(i)}, \dots, \bar{f}_{N_O^{(i)}}^{(i)}]^\top \in \mathbb{R}^{N_O^{(i)} \times D},
\end{equation}
where $\bar{f}_n^{(i)}$ denotes the preserved EMA semantic state of the $n$-th core object at the end of chunk $C^{(i)}$. In our implementation, only active entities are retained in $\mathcal{M}_{\mathrm{entity}}^{(i)}$, i.e., entities whose hidden counts do not exceed half of the maximum hidden-frame budget.
Notably, the memory footprint of $\mathcal{M}^{(i)}$ is bounded by $\mathcal{O}(D + N_O^{(i)} \cdot D)$, significantly reducing storage and computation costs during large-scale retrieval.

Our default implementation adopts a hybrid storage backend, where lightweight chunk indices remain resident on GPU for efficient retrieval, while raw visual evidence is offloaded to CPU and encoded on demand. For analysis, we additionally consider two degenerate variants: a \emph{GPU-only} backend that keeps both indices and visual evidence on GPU, and a \emph{CPU-only} backend that stores both components off-GPU. For a fair comparison under bounded device memory, the \emph{GPU-only} variant evicts the oldest visual evidence once the GPU memory budget is reached, whereas the \emph{CPU-only} variant keeps the full history off-GPU but incurs additional transfer overhead.

\begin{table*}[t]
\caption{Performance comparison on OVO-Bench and StreamingBench.}
\centering
\small
\setlength{\tabcolsep}{3.5pt}
\renewcommand{\arraystretch}{1.05}
\resizebox{\textwidth}{!}{%
\begin{tabular}{lc|ccccccc|ccccccccccc}
\toprule
\multirow{2}{*}{\textbf{Method}} & \multirow{2}{*}{\textbf{Frames}} 
& \multicolumn{7}{c|}{\textbf{OVO-Bench real-time}}
& \multicolumn{11}{c}{\textbf{StreamingBench real-time}} \\
\cmidrule(lr){3-9} \cmidrule(lr){10-20}
 &   & OCR & ACR & ATR & STU & FPD & OJR & Avg. 
 & OP & CR & CS & ATP & EU & TR & PR & SU & ACP & CT & Avg. \\
\midrule

\multicolumn{20}{l}{\textit{Proprietary Models}} \\
\midrule
Gemini 1.5 Pro ~\cite{team2024gemini} & 1 fps
& \textbf{85.9} & \underline{67.0} & \textbf{79.3} & 58.4 & 63.4 & 62.0 & 69.3 
& 79.0 & 80.5 & 83.5 & 79.7 & \textbf{80.0} & 84.7 & 77.8 & 64.2 & 72.0 & 48.7 & 75.7 \\
GPT-4o ~\cite{hurst2024gpt} & 64
& 69.8 & 64.2 & 71.6 & 51.1 & 70.3 & 59.8 & 64.5 
& 77.1 & 80.5 & 83.9 & 76.5 & 70.2 & 83.8 & 66.7 & 62.2 & 69.1 & 49.2 & 73.3 \\
\midrule
\multicolumn{20}{l}{\textit{Open-source Offline MLLMs}} \\
\midrule
LongVA ~\cite{zhang2024long} & 128
& -- & -- & -- & -- & -- & -- & --
& 70.0 & 63.3 & 61.2 & 70.9 & 62.7 & 59.5 & 61.1 & 53.7 & 54.7 & 34.7 & 60.0  \\
LongVU-7B ~\cite{shen2024longvu} & 1 fps
& 55.7 & 49.5 & 59.5 & 48.3 & 68.3 & 63.0 & 57.4
& -- & -- & -- & -- & -- & -- & -- & -- & -- & -- & -- \\
LLaVA-Video ~\cite{zhang2024llava} & 64
& 69.1 & 58.7 & 68.8 & 49.4 & 74.3 & 59.8 & 63.5
& -- & -- & -- & -- & -- & -- & -- & -- & -- & -- & -- \\

\midrule
\multicolumn{20}{l}{\textit{Open-source Online MLLMs (Training-Based)}} \\
\midrule
VideoLLM-online-8B ~\cite{chen2024videollm} & 2 fps
& 8.1 & 23.9 & 12.1 & 14.0 & 45.5 & 21.2 & 20.8 
& 39.1 & 40.1 & 34.5 & 31.1 & 46.0 & 32.4 & 31.5 & 34.2 & 42.5 & 27.9 & 36.0 \\
Dispider-7B ~\cite{qian2025dispider}& 1 fps 
& 57.7 & 49.5 & 62.1 & 44.9 & 61.4 & 51.6 & 54.6 
& 74.9 & 75.5 & 74.1 & 73.1 & 74.4 & 59.9 & 76.1 & 62.9 & 62.2 & 45.8 & 67.6 \\
Flash-VStream-7B ~\cite{zhang2025flash} & 1 fps 
& 25.5 & 32.1 & 29.3 & 33.7 & 29.7 & 28.8 & 29.9 
& 25.9 & 43.6 & 24.9 & 23.9 & 27.3 & 13.1 & 18.5 & 25.2 & 23.9 & 48.7 & 23.2 \\
ViSpeak ~\cite{fu2025vispeak}& 1 fps
& 75.2 & 58.7 & 71.6 & 51.1 & 74.3 & 66.9 & 66.3
& 79.8 & \textbf{88.3} & 83.3 & 81.1 & 76.4 & 75.1 & 70.4 & 65.9 & \underline{77.3} & 34.2 & 74.4 \\
TimeChat-Online-7B ~\cite{yao2025timechat} & 1 fps 
& 75.2 & 46.8 & 70.7 & 47.8 & 69.3 & 61.4 & 61.9 
& 80.8 & 79.7 & 80.8 & 83.3 & 74.8 & 78.8 & 78.7 & 64.2 & 68.8 & \textbf{58.0} & 75.3 \\
StreamForest-7B ~\cite{zeng2025streamforest} & 1 fps
& 68.5 & 53.2 & 71.6 & 47.8 & 65.4 & 60.9 & 61.2 
& 83.1 & \underline{82.8} & 82.7 & 84.3 & 77.5 & 78.2 & 76.9 & 69.1 & 75.6 & \underline{54.4} & 77.3 \\

\midrule
\multicolumn{20}{l}{\textit{Open-source Online MLLMs (Training-Free)}} \\
\midrule
LLaVA-OneVision-0.5B~\cite{li2024llava} &   32
 & 53.7 & 53.2 & 48.3 & 33.7 & 60.4 & 48.9& 49.7
 & 71.4 & 57.8 & 65.9 & 69.6 & 69.2 & 55.8 & 57.4 & 52.9 & 62.0 & 16.6 & 59.6 \\
\quad +ReKV~\cite{di2025streaming} &  0.5 fps
 & 41.6 & 45.0 & 50.0 & 29.8 & 60.4 & 35.9 & 43.8
& 65.1 & 60.2 & 66.6 & 66.0 & 66.7 & 53.0 & 57.4 & 48.4 & 60.3 & 18.1 & 57.4 \\
\rowcolor{blue!10}\quad +MEMO &    1 fps
& 50.3 & 56.9 & 50.9 & 37.1 & 54.5 & 52.7 & 50.4
 & 68.8 & 62.5 & 66.3 & 71.2 & 63.7 & 56.4 & 58.3 & 52.0 & 62.5 & 15.4
& 59.5  \\

\midrule

LLaVA-OneVision-7B ~\cite{li2024llava} & 32 
& 67.8 & 55.1 & 72.4 & 48.3 & 72.3 & 62.5 & 63.1  
& 80.4 & 74.2 & 76.0 & 80.7 & 72.7 & 71.7 & 67.6 & 65.5 & 65.7 & 45.1 & 71.1 \\
\quad +ReKV~\cite{di2025streaming} & 0.5 fps
& 52.4 & 54.1 & 69.8 & 43.3 & 67.3 & 57.1 & 57.3
& 74.4 & 78.9 & 78.6 & 77.1 & 68.3 & 67.9 & 67.6 & 62.6 & 64.3 & 44.6 & 69.1 \\
\quad +LiveVLM~\cite{ning2025livevlm}
& 0.5 fps 
 & -- & -- & -- & -- & -- & -- & -- 
& \underline{81.5} & 78.1 & 83.3 & 79.1 & 69.6 & 74.1 & 75.0 & 69.1 & 67.7 & 40.4 & 72.9 \\
\quad +StreamKV ~\cite{chen2026streamkv}
&  0.5 fps
 & -- & -- & -- & -- & -- & -- & --
& 73.8 & 77.3 & 85.9 & 77.5 & 73.3 & 63.9 & 69.4 & 61.4 & 63.2 & 35.8 & 68.8 \\
\rowcolor{blue!10}\quad +MEMO &    1 fps
& 71.1 & 64.2 & 75.0 & 48.3 & 74.3 & 66.9 & 66.6
& 77.7 & 69.5 & \underline{89.7} & 82.5 & 71.3 & 68.0 & 68.6 & 71.2 & 64.5  & 42.0 & 73.2 \\

\midrule
Qwen2.5-VL-7B~\cite{bai2025qwen25vltechnicalreport}
& 1 fps 
& 67.8 & 55.1 & 67.2 & 42.1 & 66.3 & 60.9 & 59.9 
& 77.9 & 76.6 & 78.6 & 80.9 & 76.7 & 77.0 & 80.6 & 65.5 & 65.7 & 52.9 & 73.3 \\
\quad +FluxMem~\cite{xie2026fluxmem} & 1 fps
& 81.2 & 59.6 & 70.7 & 53.4 & 75.2 & 63.0 & 67.2
& 80.2 & 81.1 & 81.4 & 85.3 & 78.0 & 83.8 & 80.6 & 65.9 & 69.6 & 52.1 & 76.4 \\
\rowcolor{blue!10}\quad +MEMO &   1 fps
& 75.8 & 61.5 & \underline{78.5} & 53.4 & 69.3 & 63.0 & 66.9  
& 81.0 & 78.1 & 83.9 & \underline{88.0} & 75.8 & \underline{91.0} & \textbf{86.1} & \textbf{75.2} & 70.7 & 44.7 & \underline{78.5} \\

\midrule
Qwen3-VL-8B~\cite{bai2025qwen3}  &  1 fps
& 71.1 & 65.1 & 75.9 & \underline{64.6} & \underline{75.3} & \underline{70.7} & \underline{70.1} 

& 76.8 & 77.2 & 77.3 & 80.7 & 70.4 & 75.2 & 80.6 & 64.2 & 65.8 & 49.2 & 73.2 \\

\rowcolor{blue!10}\quad +MEMO &  1 fps

& \underline{83.9} & \textbf{74.3} & \underline{78.5} & \textbf{67.4} & \textbf{76.2} & \textbf{75.5} & \textbf{76.0} 
& \textbf{89.7} & 75.8 & \textbf{90.9} & \textbf{90.3} & \underline{79.6} & \textbf{95.6} & \underline{83.3} & \underline{74.4} & \textbf{82.4} & 52.1 & \textbf{83.7}\\

\bottomrule
\end{tabular}%
}

\label{tab:benchmark-results}
\end{table*}

\subsection{Query-Specific Evidence Retrieval}
\label{sec:method_retrieval}

Given a natural-language query $Q$, we use the CLIP text Encoder to extract the text feature vector $T_Q \in \mathbb{R}^{D}$. The system first performs parallel similarity computation over all lightweight chunk indices. For each candidate chunk $C^{(i)}$, we compute the cosine similarity between $T_Q$ and its global summary $\mathcal{M}_{\mathrm{global}}^{(i)}$, as well as the best local match between $T_Q$ and its entity memory set $\mathcal{M}_{\mathrm{entity}}^{(i)}$:

\begin{equation}
\mathrm{Sim}_{\mathrm{global}}^{(i)}
=
\cos\!\big(T_Q, \mathcal{M}_{\mathrm{global}}^{(i)}\big),\quad
\mathrm{Sim}_{\mathrm{local}}^{(i)}
=
\max_{e \in \mathcal{M}_{\mathrm{entity}}^{(i)}} \cos\!\big(T_Q, e\big).
\end{equation}

The final retrieval score is obtained by linearly combining the two similarity terms:
\begin{equation}
\mathrm{Score}(Q, C^{(i)})
=
\lambda_{\mathrm{global}} \mathrm{Sim}_{\mathrm{global}}^{(i)}
+
\lambda_{\mathrm{local}} \mathrm{Sim}_{\mathrm{local}}^{(i)},
\end{equation}
where $\lambda_{\mathrm{global}}$ and $\lambda_{\mathrm{local}}$ are weighting hyperparameters.
Based on the retrieval scores, we select the Top-$K$ most relevant chunks:
\[
\mathcal{R}_K(Q)=\{C^{(i_1)}, \dots, C^{(i_K)}\}.
\]
Their associated visual evidence is denoted as:
\[
X_{\mathrm{ret}}(Q)
=
\{X^{(i)} \mid C^{(i)} \in \mathcal{R}_K(Q)\}.
\]
We further incorporate the pending frames from the current active
segment, denoted as $X_{\mathrm{cur}}$. To accommodate the bounded
visual context window of the MLLM, we uniformly sample at most
$N_{\mathrm{LLM}}$ frames from both $X_{\mathrm{ret}}(Q)$ and
$X_{\mathrm{cur}}$, yielding the sampled visual inputs
$\tilde{X}_{\mathrm{ret}}(Q)$ and $\tilde{X}_{\mathrm{cur}}$.
Only these selected frames are processed by the MLLM vision encoder.
For previously encoded evidence, the cached MLLM-specific visual
representations are directly reused. Finally, the answer $A_t$ is
generated as:
\begin{equation}
A_t =
\phi\bigl(Q,\tilde{X}_{\mathrm{ret}}(Q),\tilde{X}_{\mathrm{cur}}\bigr).
\end{equation}

\section{Experiments}
\subsection{Experimental Setup}

\textbf{Datasets.} We evaluate MEMO on two online video understanding benchmarks: the \textit{real-time} subset of StreamingBench~\cite{lin2024streamingbench} and  OVO-Bench~\cite{niu2025ovo}. StreamingBench evaluates real-time comprehension over continuous video streams, whereas OVO-Bench focuses on timestamp-anchored tasks, including historical retrieval, real-time awareness, and proactive response.

\noindent
\textbf{Baselines \& Competitors.} To verify that MEMO is training-free, plug-and-play, and compatible with diverse MLLM backbones, we evaluate it on four open-source models spanning different parameter scales and architectures, including \textit{LLaVA-OV-0.5B/7B}~\cite{li2024llava}, \textit{Qwen2.5-VL-7B}~\cite{bai2025qwen25vltechnicalreport}, and \textit{Qwen3-VL-8B}~\cite{bai2025qwen3}. For each backbone, we compare the vanilla model with its MEMO-enhanced version to isolate the performance gains brought by our framework.
In addition, we compare MEMO with other representative methods under the same evaluation protocol, grouped into four categories: (1) \textit{proprietary MLLMs}, including \textit{Gemini 1.5 Pro}~\cite{team2024gemini} and \textit{GPT-4o}~\cite{hurst2024gpt}; (2) \textit{open-source offline video MLLMs}, including \textit{LongVA}~\cite{zhang2024long}, \textit{LongVU-7B}~\cite{shen2024longvu}, and \textit{LLaVA-Video}~\cite{zhang2024llava}; (3) \textit{open-source online MLLMs with additional training}, including \textit{VideoLLM-online-8B}~\cite{chen2024videollm}, \textit{Dispider-7B}~\cite{qian2025dispider}, \textit{Flash-VStream-7B}~\cite{zhang2025flash}, \textit{ViSpeak}~\cite{fu2025vispeak}, \textit{TimeChat-Online-7B}~\cite{yao2025timechat}, and \textit{StreamForest-7B}~\cite{zeng2025streamforest}; and (4) \textit{training-free online adaptation methods}, including \textit{ReKV}~\cite{di2025streaming}, \textit{LiveVLM}~\cite{ning2025livevlm}, \textit{StreamKV}~\cite{chen2026streamkv}, \textit{Vista}~\cite{lu2026vista}, and \textit{FluxMem}~\cite{xie2026fluxmem}.

\noindent
\textbf{Implementation Details.}
As an independent pre-reasoning enhancement module, MEMO requires no parameter updates for the backbone models. For the input configuration, the incoming video stream is uniformly sampled at 1 fps, and the retrieval module retrieves the top $K=3$ chunks. For the core algorithmic hyperparameters, we empirically set $(\lambda_s,\lambda_l,\lambda_g)=(0.35,0.45,0.20)$ for online chunking and $(\lambda_{\mathrm{global}},\lambda_{\mathrm{local}})=(0.6,0.4)$ for retrieval. We set the base frame budget $N_{\mathrm{LLM}}=8$ for both the current active segment and the retrieved historical evidence, which strictly caps the total visual input at 16 frames. 
For entity perception, Grounding DINO uses bounding-box and text thresholds of 0.35 and 0.25, respectively, and cross-frame objects are associated using an IoU threshold of 0.3. We set the EMA coefficient $\rho=0.3$ and the spatial balance coefficient $\alpha=0.6$. For adaptive chunking, we use a sliding window of $n=30$ with $m=1.3$, and $(\tau_{\min},\tau_{\max})=(0.1,0.9)$. A boundary is confirmed after $N_{\mathrm{confirm}}=1$ low-similarity frame, with a minimum chunk length of $L_{\min}=4$. The hidden-frame budget is $H_{\mathrm{hid}}=5$. We keep the original video resolution without manual resizing and follow the native visual preprocessing of each backbone. Evaluations are conducted on a computing cluster equipped with NVIDIA A6000 GPUs.

\subsection{Main Results}

From the results presented in Table~\ref{tab:benchmark-results}, we draw the following observations.

\noindent
\textbf{Consistent Improvement over Base Models.}
As shown in Table~\ref{tab:benchmark-results}, MEMO improves the three
stronger backbones on both benchmarks without any parameter updates,
while maintaining comparable performance on the lightweight
LLaVA-OneVision-0.5B backbone. In particular, when integrated with
Qwen3-VL-8B, MEMO improves the OVO-Bench score from 70.1\% to 76.0\%
and the StreamingBench score from 73.2\% to 83.7\%, corresponding to
gains of 5.9 \% and 10.5 \%, respectively. The
improvements across different model families and scales demonstrate
the broad applicability of MEMO.
On StreamingBench, the gains introduced by MEMO generally increase
with backbone capability. For the LLaVA-OneVision family, MEMO slightly
decreases the average accuracy of the 0.5B model by 0.1 \%, while
improving that of the 7B model by 2.1 pts. Similarly, the improvements
increase from 5.2 \% with Qwen2.5-VL-7B to 10.5 \% with Qwen3-VL-8B.
These results suggest that stronger backbones can more effectively
exploit the query-relevant evidence provided by MEMO.

\noindent
\textbf{Significant Advantages over Existing Methods.}
With Qwen3-VL-8B as the backbone, MEMO achieves state-of-the-art performance on both benchmarks, surpassing all compared methods. MEMO shows clear advantages over existing training-free methods under the same base model, including ReKV and LiveVLM. For example, on LLaVA-OneVision-7B, MEMO exceeds ReKV by 9.3\% on OVO-Bench and 4.1\% on StreamingBench. In particular, it outperforms the previous training-based SOTA, StreamForest~\cite{zeng2025streamforest}, by 14.8\% on OVO-Bench and 6.4\% on StreamingBench. These results demonstrate that an effective memory and retrieval mechanism can make lightweight training-free methods highly competitive for streaming video understanding.

\noindent
\textbf{Fine-grained Analysis.}
A finer-grained breakdown reveals that the gains introduced by MEMO are especially pronounced on tasks that require temporally coherent event preservation and fine-grained historical evidence recall. In particular, the improvements in action perception (ACP, +16.6\%) and clip summarization (CS, +13.6\%) suggest the benefit of \textit{Online Temporal Chunking}, which better preserves semantically coherent events than fixed-length segmentation. Meanwhile, the gains in OCR (+12.8\%), object recognition (OJR, +4.8\%), and attribute perception (ATP, +9.6\%) demonstrate the value of \textit{Structured Memory Construction} and \textit{Query-Specific Evidence Retrieval}, where $\mathcal{M}_{\mathrm{global}}^{(i)}$ and $\mathcal{M}_{\mathrm{entity}}^{(i)}$ provide lightweight yet informative indices for query-relevant evidence retrieval.

\begin{figure}[t]
    \centering
    \includegraphics[width=\linewidth]{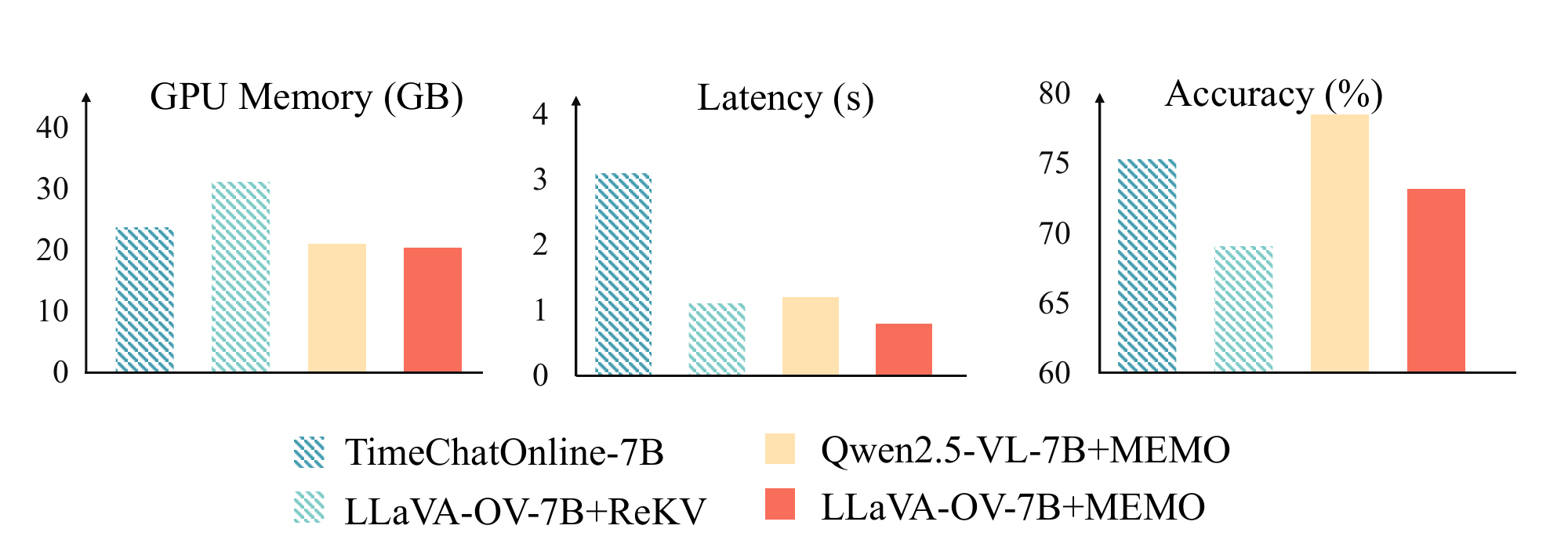}
    \caption{
    Comparison of different methods in terms of peak GPU memory usage,
    response latency, and accuracy. All experiments are conducted on a
    single NVIDIA A6000 GPU.
    }
    \Description{
    Three bar charts compare four systems in terms of peak GPU memory,
    response latency, and StreamingBench accuracy:
    TimeChat-Online-7B, LLaVA-OneVision-7B with ReKV,
    Qwen2.5-VL-7B with MEMO, and LLaVA-OneVision-7B with MEMO.
    Qwen2.5-VL-7B with MEMO achieves the highest accuracy of
    78.5 percent while using 21.1 GB of GPU memory and 1.2 seconds
    of latency. LLaVA-OneVision-7B with MEMO uses 20.5 GB and has
    the lowest latency of 0.8 seconds while reaching 73.2 percent
    accuracy. TimeChat-Online has substantially higher latency,
    whereas ReKV uses the most GPU memory and has the lowest accuracy
    among the four systems.
    }
    \label{fig:efficient}
\end{figure}

\begin{table}[t]
\centering
\caption{Ablation on perception and memory synergy.}
\label{tab:strategy_ablation}
\small
\setlength{\tabcolsep}{5pt}
\renewcommand{\arraystretch}{1.05}
\begin{tabular}{ccc|c}
\toprule
\textbf{Base} & \textbf{Perception}  & \textbf{Memory }  & \textbf{Acc. (\%)}  \\
\midrule
\cmark   & \xmark  & \xmark & 73.2 \\
\cmark & \xmark & \cmark & 79.4 \up{6.2}\\
\cmark   & \cmark  & \xmark & 77.8 \up{4.6}\\
\cmark &\cmark  & \cmark & 83.7 \up{10.5}\\
\bottomrule
\end{tabular}
\end{table}

\subsection{Efficiency Analysis}

To evaluate the inference efficiency of streaming video understanding, we benchmark the peak allocated GPU memory, end-to-end response latency, and accuracy on StreamingBench using a single NVIDIA A6000 GPU. Here, the response latency is defined as the elapsed time from receiving the input query to generating the complete textual answer, covering the entire pipeline of visual evidence retrieval, visual encoding, and MLLM decoding.

The results in Figure~\ref{fig:efficient} show that our proposed MEMO achieves a strong balance between performance and efficiency. When combined with Qwen2.5-VL-7B, MEMO attains the highest accuracy of 78.5\%, while requiring only 21.1 GB of peak memory and 1.2 seconds of average latency. Compared with TimeChat-Online, which is also built on Qwen2.5-VL-7B, MEMO improves accuracy by 3.2\% while reducing memory consumption and latency by 10.9\% and 67.6\%, respectively.
When applied to LLaVA-OneVision-7B, MEMO further reduces peak memory usage to 20.5 GB and response latency to 0.8 seconds, while still maintaining a competitive accuracy of 73.2\%, surpassing ReKV~\cite{di2025streaming} by 4.1\% under the same backbone.

To further understand the computational overhead of MEMO, we provide a
module-wise latency breakdown in Table~\ref{tab:latency_breakdown}.
The main computational cost comes from the visual perception stage, where
CLIP, Grounding DINO, and SAM require 297.60 ms per frame in total.
In contrast, the memory update and retrieval modules introduce only marginal
overhead, requiring 0.39 ms per frame and 8.08 ms per query, respectively.
These results demonstrate that MEMO's structured memory construction and
query-specific retrieval can be efficiently executed under the sampled-stream
setting.

Overall, these results demonstrate that MEMO effectively balances efficiency and accuracy in streaming video understanding. Enabled by its multi-level structured memory and query-aware retrieval mechanism, MEMO supports efficient long-context reasoning without incurring prohibitive inference overhead.

\begin{table}[t]
  \centering
  \caption{Latency breakdown of Qwen2.5-VL-7B + MEMO.
  Ret. and Gen. denote retrieval and generation, respectively.}
  \label{tab:latency_breakdown}

  \footnotesize
  \setlength{\tabcolsep}{2pt}
  \renewcommand{\arraystretch}{1.06}

  \begin{tabular*}{\columnwidth}{
    @{\extracolsep{\fill}}
    l r
    @{\hspace{5pt}}
    l r
    @{\hspace{5pt}}
    l r
    @{}
  }
    \toprule

    \multicolumn{2}{c}{\textbf{Vision / Frame}} &
    \multicolumn{2}{c}{\textbf{Memory}} &
    \multicolumn{2}{c}{\textbf{LLM / Query}} \\

    \cmidrule(lr){1-2}
    \cmidrule(lr){3-4}
    \cmidrule(lr){5-6}

    \textit{Module} & \textit{ms} &
    \textit{Operation} & \textit{ms} &
    \textit{Metric} & \textit{ms} \\
    \midrule

    CLIP
      & 16.56
      & Update
      & 0.39
      & TTFT
      & 598.44 \\

    Grounding DINO
      & 154.91
      & Ret.\ \& Recall
      & 8.08
      & TPOT
      & 55.02 \\

    SAM
      & 126.13
      &
      &
      & \textbf{Total Gen.}
      & \textbf{1367.62} \\

    \bottomrule
  \end{tabular*}

\end{table}

\subsection{Ablation Studies}

\begin{table}[t]
\centering
\small
\caption{Ablation on multi-level memory. ``Entity-only'' relies exclusively on spatial and local cues for chunking and $\mathcal{M}_{\mathrm{entity}}$ for retrieval. ``Global-only'' relies exclusively on global cues for chunking and $\mathcal{M}_{\mathrm{global}}$ for retrieval.}
\label{tab:feature_dependency}
\setlength{\tabcolsep}{5pt}
\renewcommand{\arraystretch}{1.15}
\begin{tabular}{l c c c}
\toprule
\multirow{2}{*}{\textbf{Variants}} & \multicolumn{2}{c}{\textbf{Feature Dependency}} & \multirow{2}{*}{\textbf{Acc. (\%)}} \\
\cmidrule(lr){2-3}
 & \textbf{Chunking Stage} & \textbf{Retrieval Stage} & \\
\midrule
Full Model & $S_{\mathrm{spatial}} + S_{\mathrm{local}} + S_{\mathrm{global}}$ & $\mathcal{M}_{\mathrm{entity}} + \mathcal{M}_{\mathrm{global}}$ & \textbf{83.69} \\
Entity-only & $S_{\mathrm{spatial}} + S_{\mathrm{local}}$ & $\mathcal{M}_{\mathrm{entity}}$ & 81.77 \\
Global-only & $S_{\mathrm{global}}$ & $\mathcal{M}_{\mathrm{global}}$ & 81.31 \\
\bottomrule
\end{tabular}
\end{table}

To validate the effectiveness of the key components and design choices in MEMO, we conduct comprehensive ablation studies on StreamingBench using Qwen3-VL-8B.

\noindent
\textbf{Ablation on the Synergy between Perception and Memory.} To disentangle the importance of \textit{Perception} (entity-aware perception and dynamic chunking) and \textit{Memory} (structured memory and query-specific retrieval), we construct degraded variants to isolate their individual contributions (Table~\ref{tab:strategy_ablation}). Concretely, disabling perception regresses dynamic chunking to fixed-length segmentation, while removing memory replaces retrieval with chronological truncation under the same visual budget.
The results show that memory only improves over the base model by 6.2\%, indicating that explicit memory retrieval effectively mitigates long-term forgetting. Meanwhile, perception only brings a 4.6\% gain, suggesting that adaptive perception preserves semantically coherent events more effectively than rigid segmentation. Combining both further boosts performance, revealing strong complementarity.

\begin{figure}[t]
    \centering
    \includegraphics[width=\linewidth]{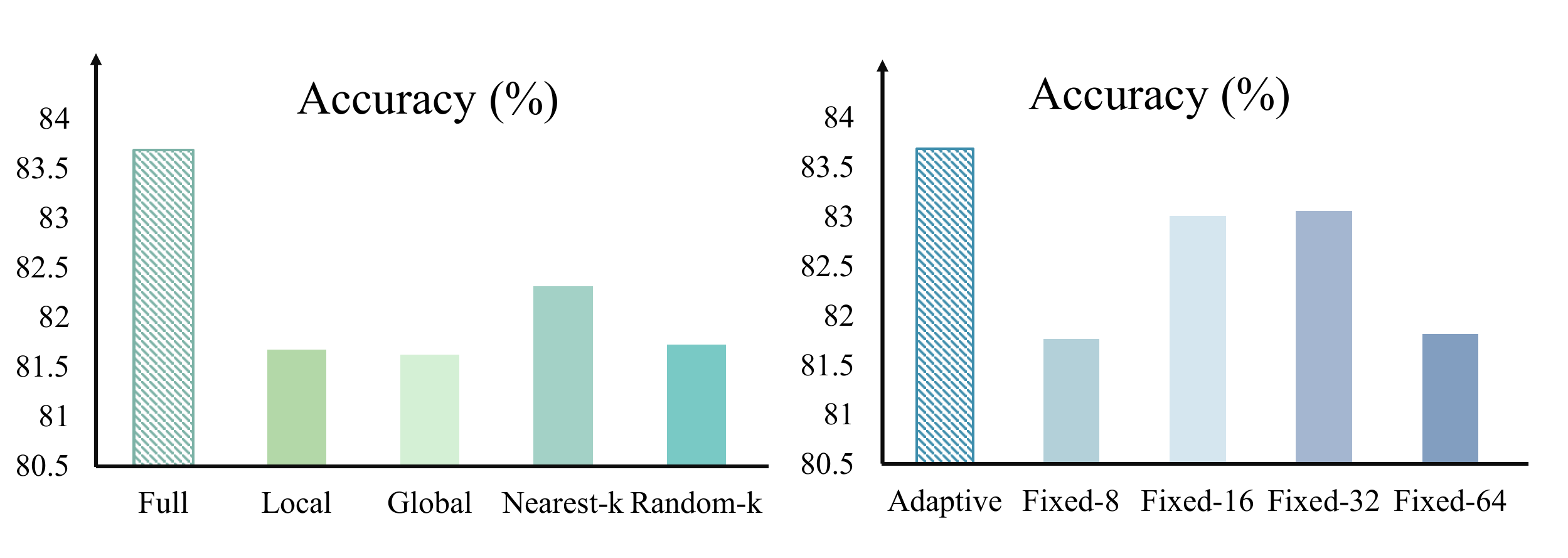}
    \caption{
    Ablation results for different retrieval strategies (left) and
    temporal chunking strategies (right).
    }
    \Description{
    Two bar charts report StreamingBench accuracy for retrieval and
    temporal chunking ablations. In the left chart, the full semantic
    retrieval strategy achieves the highest accuracy of approximately
    83.7 percent. Nearest-K retrieval ranks second at 82.32 percent,
    while the local-only, global-only, and random-K alternatives obtain
    lower scores. In the right chart, adaptive semantic chunking
    achieves the highest accuracy. Fixed-length windows of 16 and
    32 frames perform better than windows of 8 and 64 frames, but all
    fixed-length variants remain below the adaptive strategy. The
    8-frame and 64-frame settings achieve 81.77 and 81.82 percent,
    respectively.
    }
    \label{fig:ablation-retrieval-chunking}
\end{figure}

\noindent
\textbf{Ablation on Multi-Level Memory Representations.}
We evaluate the impact of hierarchical visual memory representations of the overall framework. As detailed in Table~\ref{tab:feature_dependency}, stripping the model down to an \emph{Entity-only} variant (relying exclusively on spatial/local cues for chunking and $\mathcal{M}_{\mathrm{entity}}$ for retrieval) decreases accuracy to 81.77\%, while a \emph{Global-only} configuration further drops the performance to 81.31\%. These results validate the core motivation of our multi-level design: global features capture coarse-grained scene contexts, whereas local entity features pinpoint query-relevant, object-centric evidence. Their synergistic integration is beneficial for robust streaming video understanding.

\noindent
\textbf{Ablation on Retrieval Strategies.}
Figure~\ref{fig:ablation-retrieval-chunking} (left) compares different memory retrieval strategies. The full retrieval scheme achieves the best performance, highlighting the importance of jointly exploiting local and global semantics within structured memory for accurate evidence recall.
Interestingly, the \emph{nearest-}$K$ strategy achieves competitive performance (82.32\%) by leveraging the natural temporal continuity of video streams, yet it still falls short of the full retrieval strategy. This suggests that while temporal proximity provides a valuable prior, precise memory recall ultimately demands explicit semantic matching. In contrast, \emph{random-}$K$ retrieval yields only 81.73\%, further highlighting the critical role of targeted evidence retrieval.

\noindent
\textbf{Ablation on Online Temporal Chunking.}
Figure~\ref{fig:ablation-retrieval-chunking} (right) compares our adaptive semantic chunking strategy against fixed-length chunking strategies with different window sizes. To ensure a fair comparison, the memory extraction and retrieval pipelines remain strictly identical across all configurations. The results show that all fixed-length chunking variants perform worse than our adaptive strategy. Specifically, a narrow fixed window (e.g., 8 frames, dropping to 81.77\%) tends to break semantically coherent events into fragmented pieces, disrupting the continuity of critical visual evidence. Conversely, a broad fixed window (e.g., 64 frames, 81.82\%) inevitably conflates distinct semantic events into a single chunk, diluting the relevance of the retrieved content to the user query. These findings confirm that our similarity-based dynamic chunking mechanism better aligns with the intrinsic semantic boundaries of streaming videos.

\section{Limitations}
MEMO relies on external detection, segmentation, and short-term object
association, so upstream perception errors may propagate to chunking and
retrieval. Although multi-level cues mitigate local tracking noise, the
current entity-centric indices remain less effective at representing implicit
or absence-based scene states. Persistent cross-chunk identity association
and memory consolidation for hour- or day-scale streams also remain open
problems. Future work will explore explicit scene-state memory, evidence
re-ranking, and importance-aware memory consolidation.

\section{Conclusion}
In this work, we study streaming video understanding from the perspective of memory modeling, where the key challenge lies in handling unbounded visual streams while preserving rich semantic information over time. 
We propose MEMO, a multi-level, entity-aware structured memory framework that organizes streaming video into semantically coherent chunks and represents them using both global context and fine-grained entity-level information. By modeling entity dynamics and decoupling lightweight indexing from high-resolution visual evidence, MEMO enables effective retrieval and reasoning over long video streams. Extensive experiments on StreamingBench and OVO-Bench demonstrate that MEMO consistently improves multiple base models and achieves state-of-the-art performance. We hope this work could highlight the importance of structured memory representation for streaming video understanding and also encourage future research on scalable and semantically grounded memory designs.

\begin{acks}
This work was supported by the Shanghai Municipal Science and Technology Major Project (No. 2025SHZDZX025G16).
\end{acks}

\bibliographystyle{ACM-Reference-Format}
\balance
\bibliography{acmart}

@article{di2025streaming,
  title={Streaming video question-answering with in-context video kv-cache retrieval},
  author={Di, Shangzhe and Yu, Zhelun and Zhang, Guanghao and Li, Haoyuan and Zhong, Tao and Cheng, Hao and Li, Bolin and He, Wanggui and Shu, Fangxun and Jiang, Hao},
  journal={arXiv preprint arXiv:2503.00540},
  year={2025}
}

@article{plizzari2024outlook,
  title={An Outlook into the Future of Egocentric Vision: C. Plizzari et al.},
  author={Plizzari, Chiara and Goletto, Gabriele and Furnari, Antonino and Bansal, Siddhant and Ragusa, Francesco and Farinella, Giovanni Maria and Damen, Dima and Tommasi, Tatiana},
  journal={International Journal of Computer Vision},
  volume={132},
  number={11},
  pages={4880--4936},
  year={2024},
  publisher={Springer}
}

@inproceedings{zhu2026egosound,
  title={Egosound: Benchmarking sound understanding in egocentric videos},
  author={Zhu, Bingwen and Fu, Yuqian and Dong, Qiaole and Sun, Guolei and Qian, Tianwen and Wu, Yuzheng and Paudel, Danda Pani and Fu, Yanwei and Xue, Xiangyang},
  booktitle={Proceedings of the IEEE/CVF Conference on Computer Vision and Pattern Recognition},
  pages={25589--25598},
  year={2026}
}

@inproceedings{zhang2026egonight,
  title={Egonight: Towards egocentric vision understanding at night with a challenging benchmark},
  author={Zhang, Deheng and Fu, Yuqian and Yang, Runyi and Miao, Yang and Qian, Tianwen and Zheng, Xu and Sun, Guolei and Chhatkuli, Ajad and Huang, Xuanjing and Jiang, Yu-Gang and others},
  booktitle={International Conference on Learning Representations},
  volume={2026},
  pages={887--901},
  year={2026}
}

@article{li2026bridging,
  title={Bridging the 2D-3D Gap: A Hierarchical Semantic-Geometric Map for Vision Language Navigation},
  author={Li, Kailing and Qian, Tianwen and Yang, Lijin and Fu, Yuqian and Gong, Jingyu and Wang, Xiaoling and He, Liang},
  journal={arXiv preprint arXiv:2606.00095},
  year={2026}
}

@article{brodermann2025cafuser,
  title={Cafuser: Condition-aware multimodal fusion for robust semantic perception of driving scenes},
  author={Br{\"o}dermann, Tim and Sakaridis, Christos and Fu, Yuqian and Van Gool, Luc},
  journal={IEEE Robotics and Automation Letters},
  volume={10},
  number={4},
  pages={3134--3141},
  year={2025},
  publisher={IEEE}
}

@inproceedings{wang2026streameqa,
  title={Streameqa: Towards streaming video understanding for embodied scenarios},
  author={Wang, Yifei and Li, Zhenkai and Qian, Tianwen and Zheng, Huanran and Wang, Zheng and Fu, Yuqian and Wang, Xiaoling},
  booktitle={Proceedings of the IEEE/CVF Conference on Computer Vision and Pattern Recognition},
  pages={9422--9432},
  year={2026}
}

@inproceedings{pan2026v,
  title={V2-SAM: Marrying SAM2 with Multi-Prompt Experts for Cross-View Object Correspondence},
  author={Pan, Jiancheng and Wang, Runze and Qian, Tianwen and Mahdi, Mohammad and Fu, Yanwei and Xue, Xiangyang and Huang, Xiaomeng and Van Gool, Luc and Paudel, Danda Pani and Fu, Yuqian},
  booktitle={Proceedings of the IEEE/CVF Conference on Computer Vision and Pattern Recognition},
  pages={16910--16919},
  year={2026}
}

@inproceedings{fu2025objectrelator,
  title={Objectrelator: Enabling cross-view object relation understanding across ego-centric and exo-centric perspectives},
  author={Fu, Yuqian and Wang, Runze and Ren, Bin and Sun, Guolei and Gong, Biao and Fu, Yanwei and Paudel, Danda Pani and Huang, Xuanjing and Van Gool, Luc},
  booktitle={2025 IEEE/CVF International Conference on Computer Vision (ICCV)},
  pages={6530--6540},
  year={2025},
  organization={IEEE}
}

@inproceedings{balauca2025understanding,
  title={Understanding Museum Exhibits using Vision-Language Reasoning},
  author={Balauca, Ada-Astrid and Garai, Sanjana and Balauca, Stefan and Shetty, Rasesh Udayakumar and Agrawal, Naitik and Shah, Dhwanil Subhashbhai and Fu, Yuqian and Wang, Xi and Toutanova, Kristina and Paudel, Danda Pani and others},
  booktitle={2025 IEEE/CVF International Conference on Computer Vision (ICCV)},
  pages={2227--2238},
  year={2025},
  organization={IEEE}
}

@article{dong2026objectstream,
  title={ObjectStream: Latent Objects as Memory Anchors for Streaming Video Understanding},
  author={Dong, Mingkang and Pu, Muxin and Li, Jie and Guo, Bohan and Chen, Songruo and Ren, Bin and Zheng, Xu and Zhao, Chen and Qian, Tianwen and Elhoseiny, Mohamed and others},
  journal={arXiv preprint arXiv:2607.28312},
  year={2026}
}

@inproceedings{li2026egocross,
  title={Egocross: Benchmarking multimodal large language models for cross-domain egocentric video question answering},
  author={Li, Yanjun and Fu, Yuqian and Qian, Tianwen and Xu, Qi'ao and Dai, Silong and Paudel, Danda Pani and Van Gool, Luc and Wang, Xiaoling},
  booktitle={Proceedings of the AAAI Conference on Artificial Intelligence},
  volume={40},
  number={8},
  pages={6592--6600},
  year={2026}
}

@article{ning2025livevlm,
  title={Livevlm: Efficient online video understanding via streaming-oriented kv cache and retrieval},
  author={Ning, Zhenyu and Liu, Guangda and Jin, Qihao and Ding, Wenchao and Guo, Minyi and Zhao, Jieru},
  journal={arXiv preprint arXiv:2505.15269},
  year={2025}
}

@article{yang2025streammem,
  title={Streammem: Query-agnostic kv cache memory for streaming video understanding},
  author={Yang, Yanlai and Zhao, Zhuokai and Shukla, Satya Narayan and Singh, Aashu and Mishra, Shlok Kumar and Zhang, Lizhu and Ren, Mengye},
  journal={arXiv preprint arXiv:2508.15717},
  year={2025}
}

@article{team2024gemini,
  title={Gemini 1.5: Unlocking multimodal understanding across millions of tokens of context},
  author={Team, Gemini and Georgiev, Petko and Lei, Ving Ian and Burnell, Ryan and Bai, Libin and Gulati, Anmol and Tanzer, Garrett and Vincent, Damien and Pan, Zhufeng and Wang, Shibo and others},
  journal={arXiv preprint arXiv:2403.05530},
  year={2024}
}

@article{hurst2024gpt,
  title={Gpt-4o system card},
  author={Hurst, Aaron and Lerer, Adam and Goucher, Adam P and Perelman, Adam and Ramesh, Aditya and Clark, Aidan and Ostrow, AJ and Welihinda, Akila and Hayes, Alan and Radford, Alec and others},
  journal={arXiv preprint arXiv:2410.21276},
  year={2024}
}

@inproceedings{zhang2025flash,
  title={Flash-vstream: Efficient real-time understanding for long video streams},
  author={Zhang, Haoji and Wang, Yiqin and Tang, Yansong and Liu, Yong and Feng, Jiashi and Jin, Xiaojie},
  booktitle={Proceedings of the IEEE/CVF international conference on computer vision},
  pages={21059--21069},
  year={2025}
}

@article{zhang2024long,
  title={Long context transfer from language to vision},
  author={Zhang, Peiyuan and Zhang, Kaichen and Li, Bo and Zeng, Guangtao and Yang, Jingkang and Zhang, Yuanhan and Wang, Ziyue and Tan, Haoran and Li, Chunyuan and Liu, Ziwei},
  journal={arXiv preprint arXiv:2406.16852},
  year={2024}
}

@article{shen2024longvu,
  title={Longvu: Spatiotemporal adaptive compression for long video-language understanding},
  author={Shen, Xiaoqian and Xiong, Yunyang and Zhao, Changsheng and Wu, Lemeng and Chen, Jun and Zhu, Chenchen and Liu, Zechun and Xiao, Fanyi and Varadarajan, Balakrishnan and Bordes, Florian and others},
  journal={arXiv preprint arXiv:2410.17434},
  year={2024}
}

@article{zhang2024llava,
  title={Llava-video: Video instruction tuning with synthetic data},
  author={Zhang, Yuanhan and Wu, Jinming and Li, Wei and Li, Bo and Ma, Zejun and Liu, Ziwei and Li, Chunyuan},
  journal={arXiv preprint arXiv:2410.02713},
  year={2024}
}

@inproceedings{chen2024videollm,
  title={Videollm-online: Online video large language model for streaming video},
  author={Chen, Joya and Lv, Zhaoyang and Wu, Shiwei and Lin, Kevin Qinghong and Song, Chenan and Gao, Difei and Liu, Jia-Wei and Gao, Ziteng and Mao, Dongxing and Shou, Mike Zheng},
  booktitle={Proceedings of the IEEE/CVF Conference on Computer Vision and Pattern Recognition},
  pages={18407--18418},
  year={2024}
}

@inproceedings{qian2025dispider,
  title={Dispider: Enabling video llms with active real-time interaction via disentangled perception, decision, and reaction},
  author={Qian, Rui and Ding, Shuangrui and Dong, Xiaoyi and Zhang, Pan and Zang, Yuhang and Cao, Yuhang and Lin, Dahua and Wang, Jiaqi},
  booktitle={Proceedings of the Computer Vision and Pattern Recognition Conference},
  pages={24045--24055},
  year={2025}
}

@inproceedings{fu2025vispeak,
  title={Vispeak: Visual instruction feedback in streaming videos},
  author={Fu, Shenghao and Yang, Qize and Li, Yuan-Ming and Peng, Yi-Xing and Lin, Kun-Yu and Wei, Xihan and Hu, Jian-Fang and Xie, Xiaohua and Zheng, Wei-Shi},
  booktitle={Proceedings of the IEEE/CVF International Conference on Computer Vision},
  pages={21778--21788},
  year={2025}
}

@inproceedings{yao2025timechat,
  title={Timechat-online: 80\% visual tokens are naturally redundant in streaming videos},
  author={Yao, Linli and Li, Yicheng and Wei, Yuancheng and Li, Lei and Ren, Shuhuai and Liu, Yuanxin and Ouyang, Kun and Wang, Lean and Li, Shicheng and Li, Sida and others},
  booktitle={Proceedings of the 33rd ACM International Conference on Multimedia},
  pages={10807--10816},
  year={2025}
}

@article{xu2025streamingvlm,
  title={Streamingvlm: Real-time understanding for infinite video streams},
  author={Xu, Ruyi and Xiao, Guangxuan and Chen, Yukang and He, Liuning and Peng, Kelly and Lu, Yao and Han, Song},
  journal={arXiv preprint arXiv:2510.09608},
  year={2025}
}

@article{zeng2025streamforest,
  title={Streamforest: Efficient online video understanding with persistent event memory},
  author={Zeng, Xiangyu and Qiu, Kefan and Zhang, Qingyu and Li, Xinhao and Wang, Jing and Li, Jiaxin and Yan, Ziang and Tian, Kun and Tian, Meng and Zhao, Xinhai and others},
  journal={arXiv preprint arXiv:2509.24871},
  year={2025}
}

@article{xie2026fluxmem,
  title={FluxMem: Adaptive Hierarchical Memory for Streaming Video Understanding},
  author={Xie, Yiweng and He, Bo and Wang, Junke and Zheng, Xiangyu and Ye, Ziyi and Wu, Zuxuan},
  journal={arXiv preprint arXiv:2603.02096},
  year={2026}
}

@article{li2024llava,
  title={Llava-onevision: Easy visual task transfer},
  author={Li, Bo and Zhang, Yuanhan and Guo, Dong and Zhang, Renrui and Li, Feng and Zhang, Hao and Zhang, Kaichen and Zhang, Peiyuan and Li, Yanwei and Liu, Ziwei and others},
  journal={arXiv preprint arXiv:2408.03326},
  year={2024}
}

@article{wang2025internvl3,
  title={InternVL3. 5: Advancing Open-Source Multimodal Models in Versatility},
  author={Wang, Weiyun and Gao, Zhangwei and Gu, Lixin and Pu, Hengjun and Cui, Long and Wei, X and Liu, Z and Jing, L and Ye, S and Shao, J and others},
  journal={Reasoning, and Efficiency. arXiv},
  volume={20252508},
  year={2025}
}

@article{an2025llava,
  title={Llava-onevision-1.5: Fully open framework for democratized multimodal training},
  author={An, Xiang and Xie, Yin and Yang, Kaicheng and Zhang, Wenkang and Zhao, Xiuwei and Cheng, Zheng and Wang, Yirui and Xu, Songcen and Chen, Changrui and Zhu, Didi and others},
  journal={arXiv preprint arXiv:2509.23661},
  year={2025}
}

@inproceedings{chen2026streamkv,
  title={Streamkv: Streaming video question-answering with segment-based kv cache retrieval and compression},
  author={Chen, Yilong and Bai, Xiang and Wang, Zhibin and Bai, Chengyu and Dai, Yuhan and Lu, Ming},
  booktitle={Proceedings of the AAAI Conference on Artificial Intelligence},
  volume={40},
  number={4},
  pages={3120--3128},
  year={2026}
}

@misc{liu2024llavanext,
  title={Llavanext: Improved reasoning, ocr, and world knowledge},
  author={Liu, Haotian and Li, Chunyuan and Li, Yuheng and Li, Bo and Zhang, Yuanhan and Shen, Sheng and Lee, Yong Jae},
  year={2024}
}

@inproceedings{huang2025online,
  title={Online video understanding: Ovbench and videochat-online},
  author={Huang, Zhenpeng and Li, Xinhao and Li, Jiaqi and Wang, Jing and Zeng, Xiangyu and Liang, Cheng and Wu, Tao and Chen, Xi and Li, Liang and Wang, Limin},
  booktitle={Proceedings of the Computer Vision and Pattern Recognition Conference},
  pages={3328--3338},
  year={2025}
}

@inproceedings{song2024moviechat,
  title={Moviechat: From dense token to sparse memory for long video understanding},
  author={Song, Enxin and Chai, Wenhao and Wang, Guanhong and Zhang, Yucheng and Zhou, Haoyang and Wu, Feiyang and Chi, Haozhe and Guo, Xun and Ye, Tian and Zhang, Yanting and others},
  booktitle={Proceedings of the IEEE/CVF Conference on Computer Vision and Pattern Recognition},
  pages={18221--18232},
  year={2024}
}

@inproceedings{maaz2024video,
  title={Video-chatgpt: Towards detailed video understanding via large vision and language models},
  author={Maaz, Muhammad and Rasheed, Hanoona and Khan, Salman and Khan, Fahad},
  booktitle={Proceedings of the 62nd Annual Meeting of the Association for Computational Linguistics (Volume 1: Long Papers)},
  pages={12585--12602},
  year={2024}
}

@article{bai2025qwen3,
  title={Qwen3-vl technical report},
  author={Bai, Shuai and Cai, Yuxuan and Chen, Ruizhe and Chen, Keqin and Chen, Xionghui and Cheng, Zesen and Deng, Lianghao and Ding, Wei and Gao, Chang and Ge, Chunjiang and others},
  journal={arXiv preprint arXiv:2511.21631},
  year={2025}
}

@article{liu2024streamchat,
  title={Streamchat: Chatting with streaming video},
  author={Liu, Jihao and Yu, Zhiding and Lan, Shiyi and Wang, Shihao and Fang, Rongyao and Kautz, Jan and Li, Hongsheng and Alvare, Jose M},
  journal={arXiv preprint arXiv:2412.08646},
  year={2024}
}

@article{zhang2025videollama,
  title={Videollama 3: Frontier multimodal foundation models for image and video understanding},
  author={Zhang, Boqiang and Li, Kehan and Cheng, Zesen and Hu, Zhiqiang and Yuan, Yuqian and Chen, Guanzheng and Leng, Sicong and Jiang, Yuming and Zhang, Hang and Li, Xin and others},
  journal={arXiv preprint arXiv:2501.13106},    
  year={2025}
}

@article{xiong2025streaming,
  title={Streaming video understanding and multi-round interaction with memory-enhanced knowledge},
  author={Xiong, Haomiao and Yang, Zongxin and Yu, Jiazuo and Zhuge, Yunzhi and Zhang, Lu and Zhu, Jiawen and Lu, Huchuan},
  journal={arXiv preprint arXiv:2501.13468},
  year={2025}
}

@inproceedings{lu2026vista,
  title={Vista: Scene-Aware Optimization for Streaming Video Question Answering Under Post-Hoc Queries},
  author={Lu, Haocheng and Zhang, Nan and Tao, Wei and Qu, Xiaoyang and Li, Guokuan and Wan, Jiguang and Wang, Jianzong},
  booktitle={Proceedings of the AAAI Conference on Artificial Intelligence},
  volume={40},
  number={9},
  pages={7539--7547},
  year={2026}
}

@article{wang2022omnivl,
  title={Omnivl: One foundation model for image-language and video-language tasks},
  author={Wang, Junke and Chen, Dongdong and Wu, Zuxuan and Luo, Chong and Zhou, Luowei and Zhao, Yucheng and Xie, Yujia and Liu, Ce and Jiang, Yu-Gang and Yuan, Lu},
  journal={Advances in neural information processing systems},
  volume={35},
  pages={5696--5710},
  year={2022}
}

@misc{bai2025qwen25vltechnicalreport,
      title={Qwen2.5-VL Technical Report}, 
      author={Shuai Bai and Keqin Chen and Xuejing Liu and Jialin Wang and Wenbin Ge and Sibo Song and Kai Dang and Peng Wang and Shijie Wang and Jun Tang and Humen Zhong and Yuanzhi Zhu and Mingkun Yang and Zhaohai Li and Jianqiang Wan and Pengfei Wang and Wei Ding and Zheren Fu and Yiheng Xu and Jiabo Ye and Xi Zhang and Tianbao Xie and Zesen Cheng and Hang Zhang and Zhibo Yang and Haiyang Xu and Junyang Lin},
      year={2025},
      eprint={2502.13923},
      archivePrefix={arXiv},
      primaryClass={cs.CV},
      url={https://arxiv.org/abs/2502.13923}, 
}

@article{oquab2023dinov2,
  title={Dinov2: Learning robust visual features without supervision},
  author={Oquab, Maxime and Darcet, Timoth{\'e}e and Moutakanni, Th{\'e}o and Vo, Huy and Szafraniec, Marc and Khalidov, Vasil and Fernandez, Pierre and Haziza, Daniel and Massa, Francisco and El-Nouby, Alaaeldin and others},
  journal={arXiv preprint arXiv:2304.07193},
  year={2023}
}

@article{ravi2024sam,
  title={Sam 2: Segment anything in images and videos},
  author={Ravi, Nikhila and Gabeur, Valentin and Hu, Yuan-Ting and Hu, Ronghang and Ryali, Chaitanya and Ma, Tengyu and Khedr, Haitham and R{\"a}dle, Roman and Rolland, Chloe and Gustafson, Laura and others},
  journal={arXiv preprint arXiv:2408.00714},
  year={2024}
}

@article{li2025videochat,
  title={Videochat: Chat-centric video understanding},
  author={Li, KunChang and He, Yinan and Wang, Yi and Li, Yizhuo and Wang, Wenhai and Luo, Ping and Wang, Yali and Wang, Limin and Qiao, Yu},
  journal={Science China Information Sciences},
  volume={68},
  number={10},
  pages={200102},
  year={2025},
  publisher={Springer}
}

@article{chen2024end,
  title={End-to-end autonomous driving: Challenges and frontiers},
  author={Chen, Li and Wu, Penghao and Chitta, Kashyap and Jaeger, Bernhard and Geiger, Andreas and Li, Hongyang},
  journal={IEEE Transactions on Pattern Analysis and Machine Intelligence},
  volume={46},
  number={12},
  pages={10164--10183},
  year={2024},
  publisher={IEEE}
}

@inproceedings{lei2021less,
  title={Less is more: Clipbert for video-and-language learning via sparse sampling},
  author={Lei, Jie and Li, Linjie and Zhou, Luowei and Gan, Zhe and Berg, Tamara L and Bansal, Mohit and Liu, Jingjing},
  booktitle={Proceedings of the IEEE/CVF conference on computer vision and pattern recognition},
  pages={7331--7341},
  year={2021}
}

@article{shao2025tokens,
  title={When tokens talk too much: A survey of multimodal long-context token compression across images, videos, and audios},
  author={Shao, Kele and Tao, Keda and Zhang, Kejia and Feng, Sicheng and Cai, Mu and Shang, Yuzhang and You, Haoxuan and Qin, Can and Sui, Yang and Wang, Huan},
  journal={arXiv preprint arXiv:2507.20198},
  year={2025}
}

@inproceedings{zheng2024towards,
  title={Towards learning a generalist model for embodied navigation},
  author={Zheng, Duo and Huang, Shijia and Zhao, Lin and Zhong, Yiwu and Wang, Liwei},
  booktitle={Proceedings of the IEEE/CVF Conference on Computer Vision and Pattern Recognition},
  pages={13624--13634},
  year={2024}
}

@article{yang2025livestar,
  title={LiveStar: Live Streaming Assistant for Real-World Online Video Understanding},
  author={Yang, Zhenyu and Zhang, Kairui and Hu, Yuhang and Wang, Bing and Qian, Shengsheng and Wen, Bin and Yang, Fan and Gao, Tingting and Dong, Weiming and Xu, Changsheng},
  journal={arXiv preprint arXiv:2511.05299},
  year={2025}
}

@article{liu2023visual,
  title={Visual instruction tuning},
  author={Liu, Haotian and Li, Chunyuan and Wu, Qingyang and Lee, Yong Jae},
  journal={Advances in neural information processing systems},
  volume={36},
  pages={34892--34916},
  year={2023}
}

@article{qian2024streaming,
  title={Streaming long video understanding with large language models},
  author={Qian, Rui and Dong, Xiaoyi and Zhang, Pan and Zang, Yuhang and Ding, Shuangrui and Lin, Dahua and Wang, Jiaqi},
  journal={Advances in Neural Information Processing Systems},
  volume={37},
  pages={119336--119360},
  year={2024}
}

@inproceedings{radford2021learning,
  title={Learning transferable visual models from natural language supervision},
  author={Radford, Alec and Kim, Jong Wook and Hallacy, Chris and Ramesh, Aditya and Goh, Gabriel and Agarwal, Sandhini and Sastry, Girish and Askell, Amanda and Mishkin, Pamela and Clark, Jack and others},
  booktitle={International conference on machine learning},
  pages={8748--8763},
  year={2021},
  organization={PmLR}
}

@inproceedings{liu2024grounding,
  title={Grounding dino: Marrying dino with grounded pre-training for open-set object detection},
  author={Liu, Shilong and Zeng, Zhaoyang and Ren, Tianhe and Li, Feng and Zhang, Hao and Yang, Jie and Jiang, Qing and Li, Chunyuan and Yang, Jianwei and Su, Hang and others},
  booktitle={European conference on computer vision},
  pages={38--55},
  year={2024},
  organization={Springer}
}

@inproceedings{kirillov2023segment,
  title={Segment anything},
  author={Kirillov, Alexander and Mintun, Eric and Ravi, Nikhila and Mao, Hanzi and Rolland, Chloe and Gustafson, Laura and Xiao, Tete and Whitehead, Spencer and Berg, Alexander C and Lo, Wan-Yen and others},
  booktitle={Proceedings of the IEEE/CVF international conference on computer vision},
  pages={4015--4026},
  year={2023}
}

@article{lin2024streamingbench,
  title={Streamingbench: Assessing the gap for mllms to achieve streaming video understanding},
  author={Lin, Junming and Fang, Zheng and Chen, Chi and Wan, Zihao and Luo, Fuwen and Li, Peng and Liu, Yang and Sun, Maosong},
  journal={arXiv preprint arXiv:2411.03628},
  year={2024}
}

@inproceedings{niu2025ovo,
  title={Ovo-bench: How far is your video-llms from real-world online video understanding?},
  author={Niu, Junbo and Li, Yifei and Miao, Ziyang and Ge, Chunjiang and Zhou, Yuanhang and He, Qihao and Dong, Xiaoyi and Duan, Haodong and Ding, Shuangrui and Qian, Rui and others},
  booktitle={Proceedings of the Computer Vision and Pattern Recognition Conference},
  pages={18902--18913},
  year={2025}
}

\end{document}